\documentclass[sigconf]{acmart}
\AtBeginDocument{%
  }

\copyrightyear{2026}
\acmYear{2026}
\setcopyright{cc}
\setcctype{by}
\acmConference[HRI '26]{Proceedings of the 21st ACM/IEEE International Conference on Human-Robot Interaction}{March 16--19, 2026}{Edinburgh, Scotland Uk}
\acmBooktitle{Proceedings of the 21st ACM/IEEE International Conference on Human-Robot Interaction (HRI '26), March 16--19, 2026, Edinburgh, Scotland Uk}
\acmPrice{}
\acmDOI{10.1145/3757279.3785588}
\acmISBN{979-8-4007-2128-1/2026/03}

\usepackage{booktabs,makecell,threeparttable,xcolor,microtype}
\definecolor{tableHeader}{RGB}{245,245,248}
\definecolor{tableShade}{RGB}{251,251,253}

\newenvironment{tighttable}{
  \begingroup
  \setlength{\tabcolsep}{4.5pt}
  
  \small
}{\endgroup}

\usepackage{tikz}
\usetikzlibrary{arrows.meta,positioning,shapes,fit,bending}

\begin{document}

\title{From Metrics to Meaning: Insights from a Mixed-Methods Field Experiment on Retail Robot Deployment}

\author{Sichao Song}
\affiliation{%
  \institution{CyberAgent}
  \city{Tokyo}
  \country{japan}}
\additionalaffiliation{%
  \institution{The University of Osaka}
  \city{Osaka}
  \country{Japan}
}
\email{song\_sichao@cyberagent.co.jp}

\author{Yuki Okafuji}
\authornotemark[1]
\affiliation{%
  \institution{CyberAgent}
  \city{Tokyo}
  \country{japan}}
\email{okafuji\_yuki\_xd@cyberagent.co.jp}

\author{Takuya Iwamoto}
\authornotemark[1]
\affiliation{%
  \institution{CyberAgent}
  \city{Tokyo}
  \country{japan}}
\email{iwamoto\_takuya\_xa@cyberagent.co.jp}

\author{Jun Baba}
\authornotemark[1]
\affiliation{%
  \institution{CyberAgent}
  \city{Tokyo}
  \country{japan}}
\email{baba\_jun@cyberagent.co.jp}

\author{Hiroshi Ishiguro}
\affiliation{%
  \institution{The University of Osaka}
  \city{Toyonaka}
  \country{japan}}
\email{ishiguro@irl.sys.es.osaka-u.ac.jp}

\renewcommand{\shortauthors}{Sichao et al.}

\begin{abstract}

We report a mixed-methods field experiment of a conversational service robot deployed under everyday staffing discretion in a live bedding store. Over 12 days we alternated three conditions--Baseline (no robot), Robot-only, and Robot+Fixture--and video-annotated the service funnel from passersby to purchase. An explanatory sequential design then used six post-experiment staff interviews to interpret the quantitative patterns.

Quantitatively, the robot increased stopping per passerby (highest with the fixture), yet clerk-led downstream steps per stopper--clerk approach, store entry, assisted experience, and purchase--decreased. Interviews explained this divergence: clerks avoided interrupting ongoing robot-customer talk, struggled with ambiguous timing amid conversational latency, and noted child-centered attraction that often satisfied curiosity at the doorway. The fixture amplified visibility but also anchored encounters at the threshold, creating a well-defined micro-space where needs could ``close'' without moving inside.

We synthesize these strands into an integrative account from the initial show of interest on the part of a customer to their entering the store and derive actionable guidance. The results advance the understanding of interactions between customers, staff members, and the robot and offer practical recommendations for deploying service robots in high-touch retail. 
% We synthesize these strands into an integrative account--\emph{Attract} $\rightarrow$ \emph{Handoff} $\rightarrow$ \emph{Interior}--and derive actionable guidance. The results advance understanding of customer-robot-clerk coordination and offer practical recommendations for deploying service robots in high-touch retail.

\end{abstract}

\begin{CCSXML}
<ccs2012>
   <concept>
       <concept_id>10003120.10003121.10011748</concept_id>
       <concept_desc>Human-centered computing~Empirical studies in HCI</concept_desc>
       <concept_significance>300</concept_significance>
       </concept>
 </ccs2012>
\end{CCSXML}

\ccsdesc[300]{Human-centered computing~Empirical studies in HCI}

\keywords{Human-Robot Interaction (HRI), Retail Robot, Mixed-Methods, Field Experiment, Staff Perspective, Service Robots}
\begin{teaserfigure}
  \centering
  \includegraphics[width=0.75\textwidth]{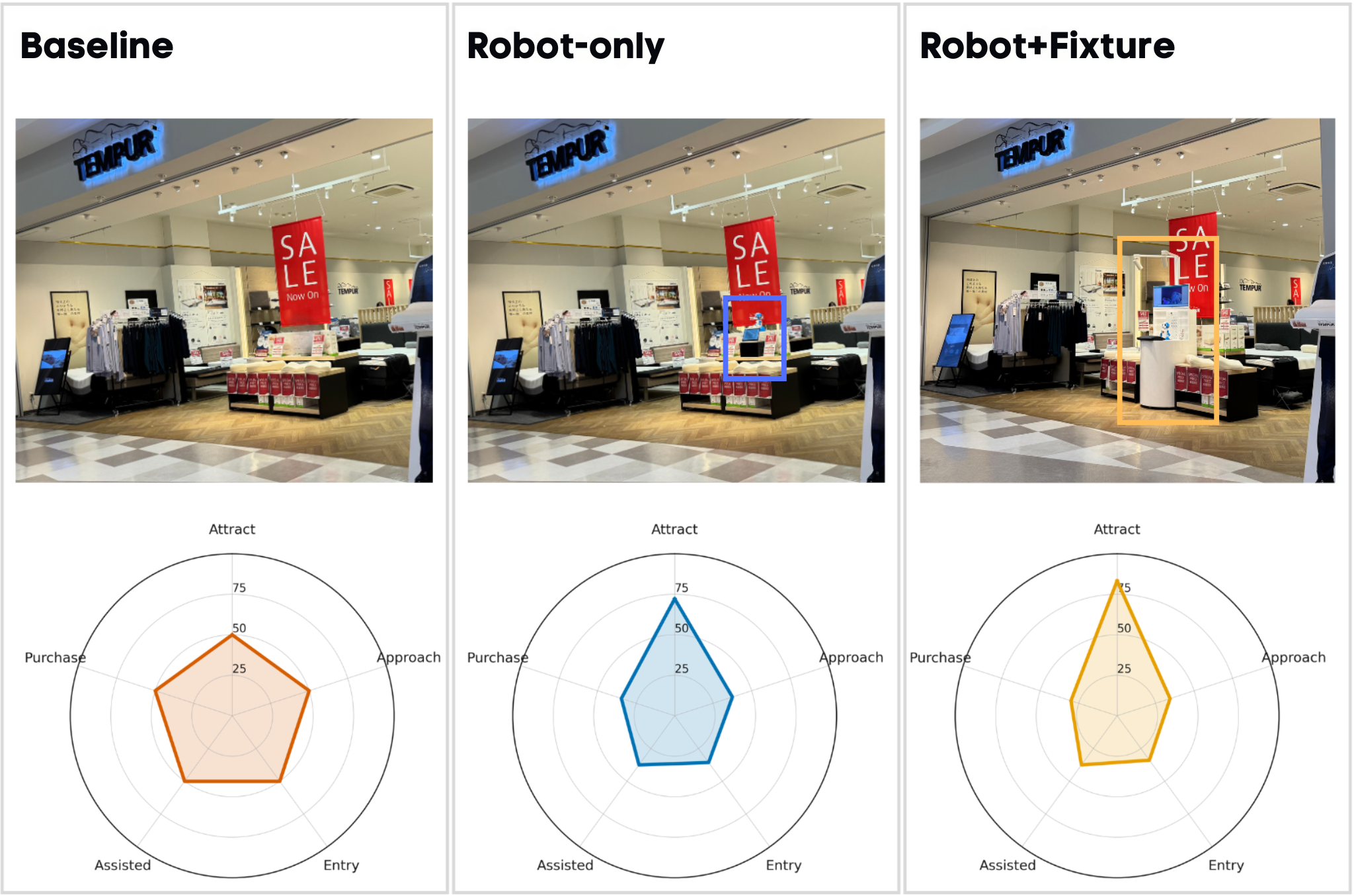}
  \caption{Top: storefront scenes for Baseline, Robot-only, and Robot+Fixture. Bottom: radar plots for Attract, Approach, Entry, Assisted, Purchase (0–100 scale with Baseline fixed at 50 on each axis; Purchase is computed per stopper). The robot increases Attract/Stopping (highest with the fixture), yet clerk-led steps (Approach, Entry, Assisted) decline, leaving purchase per stopper roughly flat.}
  \Description{Two photos depicting the entrance staging with a robot alone and with a fixture.}
  \label{fig:experimentconditions}
\end{teaserfigure}



\maketitle

\section{Introduction}

Service robots are increasingly deployed in retail to capture the attention of passersby \cite{Song2021TeleopSales,iwamoto2022pick,saad2019welcoming}, promote products \cite{Song2022BakeryField,hatano2024field,matsumura2022animation}, stage brand experiences \cite{savery2024long,chan2019examining,hwang2023relationships}, and support frontline work \cite{paluch2022my,willems2023frontline,wirtz2018brave}. Although still relatively rare, field studies consistently report increases in attention, stopping, and dwell time \cite{Kanda2009AffectiveHRI,Okafuji2022MallBehavior,Brengman2021,DeGauquierIn2021,Kanda2010ShoppingMallTRO}, and in some settings, shifts in downstream behaviors such as trials, recommendations, and purchases \cite{Song2023OutForInHRI,Song2022BakeryField,watanabe2018department,graef2023buy,SongWingman2024}. Yet converting attraction into meaningful engagement is nontrivial: outcomes hinge on how deployments are staged at the entrance, how space and talk are coordinated, and whether customers perceive the setting as approachable \cite{OH201227,DeGauquierIn2021,RobotStore1982}. To address these questions, prior work has mapped key in-store metrics---passersby, stops, touches, and purchases---and examined how effects vary with different placements and roles of a robot \cite{Shiomi2017Invite,Brengman2021,DeGauquierIn2021,SongFrom2025}.

Despite this progress, much of the literature remains metrics-centric and short-term, often focusing on a single stakeholder (the customer) and offering limited integration of frontline staff interpretations within the same deployment. This is problematic because interaction dynamics in real stores are triadic: \emph{robot-clerk-customer} coordination determines whether attention translates into approach and assistance \cite{Niemela2017PepperMall,DeGauquier2023Team,Belanche11032020,OkafujiDesign2025}. Accordingly, recent studies have begun to examine clerk-robot collaboration and how to make it explicit in carrying customers from threshold attention to interior engagement \cite{Okafuji2023Frontiers,SongFrom2025,LiRobotics2021}. Long-term fieldwork complements these insights by showing how customer behavior and clerk expectations evolve in practice and what ``usefulness'' comes to mean for both sides \cite{Song2024NewComer}. However, studies that take a holistic, multi-stakeholder view within the same live deployment remain rare. Without such work, practical know-how remains out of reach.

From another perspective, physical retail stores are increasingly designed as staged experiences, not just points of sale. Store atmospherics---layout, signage, fixtures, and how staff appear---shape attention and approach decisions before any conversation begins \cite{RobotStore1982,Bitner1992Servicescapes,OtterbringDecompression2018}. In this lens, a robot alone is insufficient. On its own, a robot mostly boosts stopping and curiosity at the entrance, but its downstream impact can be limited. In this study, we considered whether meaningful conversion would emerge when a robot was staged in a micro-boutique, that is, a small, curated entrance scene anchored by a fixture that makes a legible offer, signals what to do next, and prepares customers for a smooth handoff to a human sales representative. Such pragmatic staging elements have rarely been examined in prior research.

Motivated by these issues, we conducted a mixed-methods field experiment in a bedding retail store in Japan (Fig.~\ref{fig:experimentconditions}). In partnership with the store manager, we refined the concept, summarized design principles, and established research goals. Over a 12-day in-store deployment, we compared three conditions---Baseline (no robot), Robot-only, and Robot+Fixture---while video-annotating customer behavior and downstream outcomes (e.g., clerk approach, store entry, purchase). By incorporating experimental conditions using a fixture, we simulated the introduction of a robot in a manner that is closer to an actual implementation rather than simply setting up a robot as a demonstration. During the study, we focused not only on the robot’s impact but also examined whether and how the fixture could serve as a feasible and effective staging choice for the robot. We then conducted post-experiment interviews with six clerks to interpret the observed patterns and look deeply into how clerks made sense of working together with the robot. Our findings connect \emph{what changed} to \emph{why it changed}, surfacing interpretations, strategies, and frictions behind the metrics.

\subsection{Research Objectives (ROs)}

To orient the study, we articulate two research objectives that connect quantitative outcomes with frontline interpretations:

\begin{itemize}
\item \textbf{RO1:} To investigate how the deployment of a service robot, including the use of a fixture as a staging choice, in a real retail environment affects customer behavior and service outcomes.
\item \textbf{RO2:} To understand how retail staff perceive and interpret the effects of robot deployment, and to identify emerging considerations for practical implementation.
\end{itemize}

For RO1, we quantify differences across conditions in stopping rate (per passerby), clerk approach, store entry, and purchase outcomes (per stopper). For RO2, we analyze staff narratives about intervention timing, role allocation among clerks, the robot, and the fixture, and perceived trade-offs.

\subsection{Contributions}

This paper contributes to HRI in three ways. First, we provide field evidence from a three-condition deployment linking robot presence to attention and downstream service outcomes in a live store. Second, we complement metrics with frontline interpretations that explain counterintuitive patterns (e.g., attention up with fewer clerk approaches) via non-intrusion norms, entrance anchoring, and role negotiation. Third, we distill actionable recommendations for deployment, including explicit yet lightweight handoffs, spatial separation of ``attract'' vs.\ ``recommend,'' and staging choices that preserve approachability while leveraging attention gains.

\section{Methodology}

This study was conducted in collaboration with a bedding retail store during normal business operations. The deployment and data handling followed institutional ethical review and a store agreement. The study was approved by the Research Ethics Committee of The University of Osaka (Reference Number: R1–5–9).

\subsection{Mixed-Methods Approach}

We adopted an explanatory sequential mixed-methods design \cite{ivankova2006using,creswell2017designing}. During the field experiment, two researchers were on site, maintained structured observation logs, and captured preliminary counts (e.g., entry and clerk approach) to surface salient phenomena in real time. The store also operated an in-house counter for store entries, which we accessed during the deployment. In the week immediately after the deployment, we conducted semi-structured interviews with clerks while experiences were fresh; prompts were informed by on-site observations and preliminary indicators, preserving the explanatory intent. After interviews, we completed detailed video annotation and statistical analysis, then integrated the quantitative and qualitative strands at interpretation to connect observed patterns to clerk sense-making.

\begin{figure}[!t]
  \centering
  \includegraphics[width=0.7\linewidth]{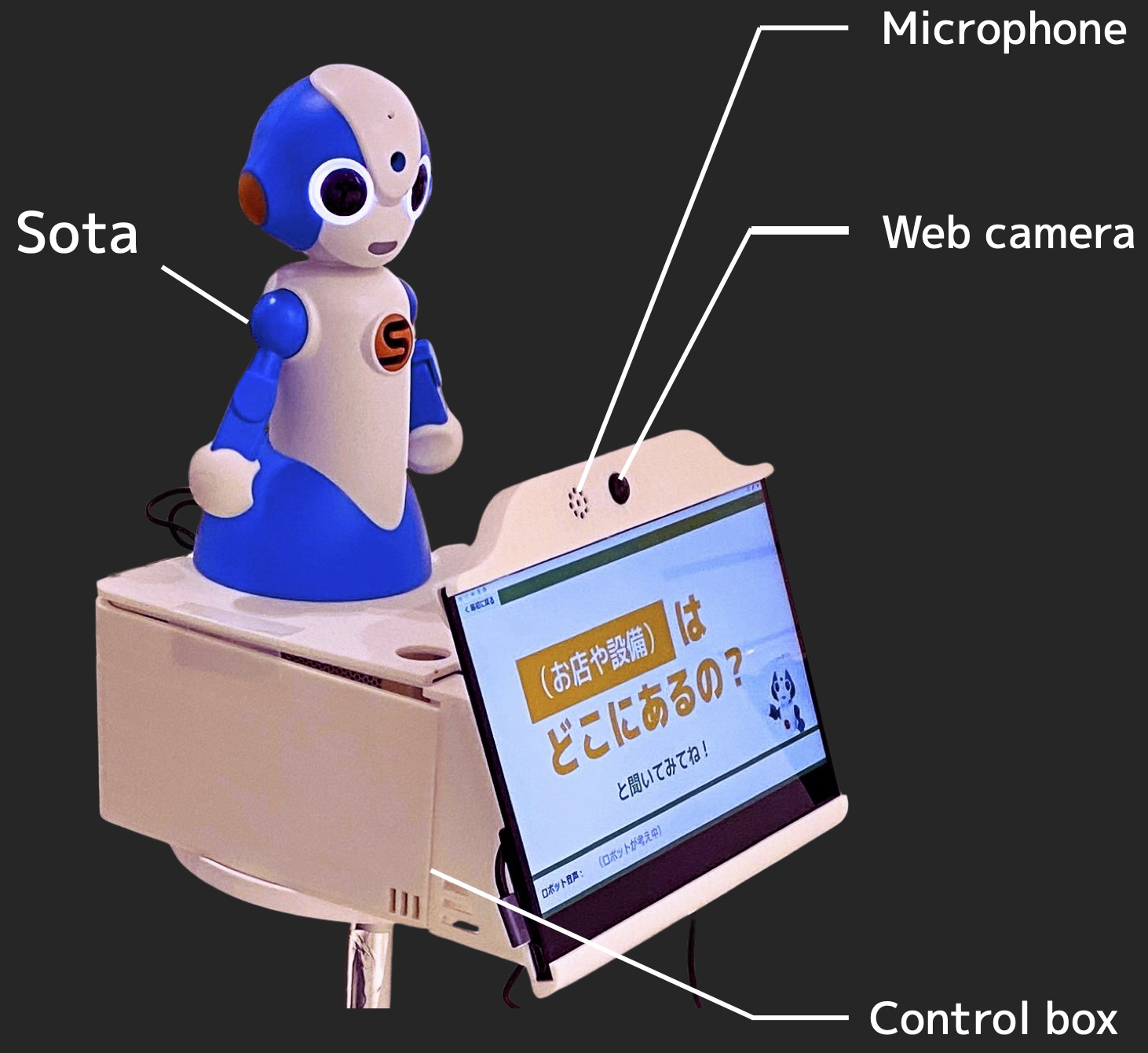}
  \caption{Service robot system (robot, base, and display).}
  \Description{Service robot system.}
  \label{fig:robotsystemlatest}
\end{figure}

\subsection{Service Robot System}

We used a small humanoid robot (``Sota,'' Vstone Co., Ltd.; approx.\ 28\,cm tall) with a childlike appearance. The robot performs simple gestures with its arms and body rotation and features three facial LEDs (eyes and mouth) that support expressive cues. As shown in Fig.~\ref{fig:robotsystemlatest}, a compact control box under the robot houses a mini PC that runs a behavior controller. For speech, we used Google Cloud Speech-to-Text and Google Cloud Text-to-Speech APIs. For conversational content generation and dialogue-state management, we used the OpenAI API (GPT-4 family). The front display presented short prompts, product information, and simple calls to action.

\subsection{Fixture}

\begin{figure}[!t]
  \centering
  \includegraphics[width=0.85\linewidth]{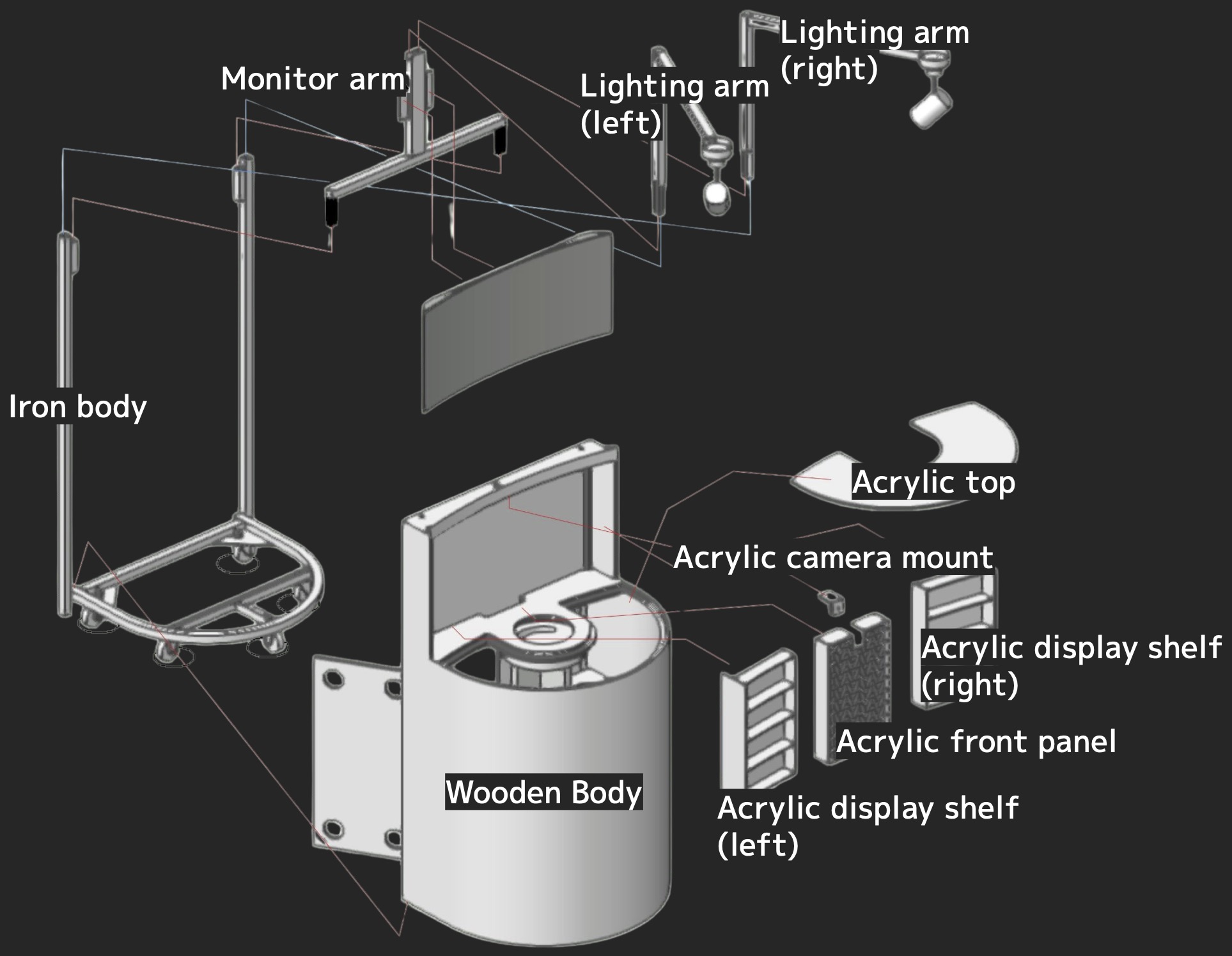}
  \caption{Fixture used for entrance staging (robot platform, acrylic top for sample, and storage).}
  \Description{Fixture used for entrance staging.}
  \label{fig:fixturedesign}
\end{figure}

We co-designed a simple entrance fixture with the store (Fig.~\ref{fig:fixturedesign}). The unit provided a stable platform for the robot and an acrylic top to place a sample pillow in front of the robot for immediate touch. Dimensions were approximately 0.70\,m (W) $\times$ 0.58\,m (D) $\times$ 1.68\,m (H); the platform height was about 0.82\,m, enabling comfortable eye height for most shoppers. The finish was neutral (white) to match store aesthetics; lighting was considered to improve visibility without cluttering the doorway.

\subsection{Field Experiment Design}

\textbf{Store.}
% \paragraph{\textbf{Store.}}
The site was a bedding retail outlet specializing in mattresses, pillows, and sleep accessories. Clerk hospitality is central to the service model. Prior to the study, we held design meetings with the store manager to agree on the robot’s role and utterance policy, staff coordination policy, and data we could collect during business operations. During the study, sample pillows were placed near the robot so that customers could physically engage while interacting.

\textbf{Conditions.}
% \paragraph{\textbf{Conditions.}}
We compared three conditions:
(1) \emph{Baseline}: normal operation with no robot system (see Fig.~\ref{fig:experimentconditions}, Baseline;
(2) \emph{Robot-only}: a conversational robot at the entrance delivering brief greetings and product prompts on a fixed schedule (see Fig.~\ref{fig:experimentconditions}, Robot-only);
(3) \emph{Robot+Fixture}: the same robot mounted on the entrance fixture (Fig.~\ref{fig:experimentconditions}, Robot+Fixture).

\textbf{Duration.}
% \paragraph{\textbf{Duration.}}
We secured 12 days in total (4 days per condition). Conditions alternated across days to balance weekday/weekend traffic. A storewide promotion was running throughout the period with consistent content, reducing the likelihood of short-term shocks.

\textbf{Operational policy.}
% \paragraph{\textbf{Operational policy.}}
Clerks were not required to approach customers. To keep the setting natural, staff used their own judgment within normal work constraints. We did not script handoffs, as this would add operational burden and could conflict with other duties and local norms.

\textbf{Measures.}
% \paragraph{\textbf{Measures.}}
Two cameras captured activity: a webcam mounted on the robot system (for entrance interactions) and a 360° camera positioned inside the store (for wider context). Two researchers kept an observation log. We report \emph{Stopping} per passerby and downstream outcomes per stopper: \emph{clerk approach}, \emph{store entry}, \emph{staff-led entry}, \emph{assisted in-store experience}, \emph{purchase}, and \emph{sample touch near the robot}. Particularly, the focus on percentages per stopper was intended to capture conversion among customers clearly influenced by the robot.
%We report \emph{Stopping} per passerby (i.e., the fraction of passersby who paused and oriented to the entrance setup) and downstream outcomes \emph{per stopper}: \emph{clerk approach}, \emph{store entry}, \emph{staff-led entry} (approach followed by entry within a short window), \emph{assisted in-store experience} (guided tryout/explanation), \emph{purchase}, and \emph{sample touch near the robot}. Particularly, the focus on percentages per stopper was intended to capture conversion among customers clearly influenced by the robot.

\textbf{Analysis.}
% \paragraph{\textbf{Analysis.}}
We used Pearson Chi-square tests of independence (reporting $\chi^2$, df, and $p$). Significant omnibus tests triggered Bonferroni-adjusted pairwise chi-square tests. Stopping is reported per passerby; all downstream outcomes are reported per stopper. Quantitative results are later integrated with qualitative themes.

\subsection{Post-Experiment Interview Study}

\textbf{Participants.}
% \paragraph{\textbf{Participants.}}
Six employees who worked during the deployment participated: one store manager (male, 40s) and five clerks (female, 50s; male, 50s; female, 60s; male, 20s; female, 20s). All of them had almost no prior experience with robots.
Sessions were conducted in the week immediately after the field experiment to minimize recall bias and align prompts with phenomena observed on site.

\textbf{Protocol.}
% \paragraph{\textbf{Protocol.}}
Semi-structured interviews (30--60 minutes) covered perceived customer reactions, intervention timing, role allocation among clerks/robot/fixture, perceived trade-offs (approachability vs.\ disruption), and operational considerations (e.g., workload, peak hours). All sessions were conducted on site in a private staff room to ensure privacy. Audio was recorded (with consent) using Zoom; automatic transcripts were exported and then checked and corrected by a native Japanese speaker before analysis. Interviews were conducted in Japanese by the first author (lead interviewer) with the second author as an assistant interviewer who took notes and asked occasional follow-ups. The interview guide was informed by on-site observation logs and preliminary counts gathered during the deployment so as to probe salient phenomena while memories were fresh.

\textbf{Interview prompts.}
% \paragraph{\textbf{Interview prompts.}}
Typical prompts included:
\begin{itemize}
  \item Ice-breakers on role, shift patterns, and IT literacy.
  \item Daily tasks and usual service scripts; how clerks normally engage customers.
  \item Perceived changes in tasks, motivation, and communication styles after robot deployment (Robot-only / Robot+Fixture).
  \item Situations that prompted or delayed intervention (e.g., avoiding interruption while the robot talked); cues that would make handoffs easier.
  \item Perceived effects by customer segment (e.g., families with children vs.\ solo customers) and by product type (e.g., higher-priced bedding).
  \item Strengths, weaknesses, and suggestions for improvement in staging, layout, and robot utterances.
\end{itemize}

We also used targeted probes to follow up on phenomena observed during the experiment (e.g., entrance anchoring, timing of clerk approach), and we invited concrete episodes (what happened, who did what, what was said) to ground interpretations.

\textbf{Analysis.}
% \paragraph{\textbf{Analysis.}}
We conducted a reflexive thematic analysis \cite{VirginiaUsing2006} following an iterative, inductive, and interpretive process supported by shared Google Spreadsheets for collaborative coding and memoing \cite{OgawaUnderstanding2025}. Specifically, we proceeded as follows:
\begin{enumerate}
  \item Familiarization: the first and second authors read the transcripts and wrote analytic memos about salient moments.
  \item Initial coding: the first author coded each transcript using labels and brief summaries of properties.
  \item Code discussion and consolidation: after each transcript, the first and second authors reviewed codes, refined labels, and noted preliminary theme candidates.
  \item Theme development: codes were grouped, compared, and iteratively reorganized into candidate themes; we actively sought deviant cases that challenged emerging interpretations.
  \item Review and definition: themes were reviewed against the dataset and clarified in relation to the research objectives, with names and scope refined.
  \item Reporting: representative, de-identified excerpts were selected to illustrate each theme and linked to quantitative funnel outcomes where relevant.
\end{enumerate}
To enhance rigor, we maintained a decision log and analytic memos throughout, and used peer debriefing meetings to challenge early interpretations.

\subsection{Data Handling}
All data practices followed the approved protocol and store agreement. Experimental video files were stored on encrypted drives. We annotated videos locally and exported only derived tables for analysis. Interview audio was recorded with consent, transcribed verbatim, and pseudonymized (e.g., P1--P6). Reporting focuses on aggregated statistics and de-identified excerpts; potentially identifying details were paraphrased, and any illustrative images are cropped or masked when necessary.

\begin{table*}[t]
\centering
\caption{Metric analysis with trend tests. Stopping is per passerby; others per stopper.}
\label{tab:quant_summary}
\begin{threeparttable}
\begin{tighttable}
\begin{tabular}{@{}l c c c c c c@{}}
\toprule
\textbf{Metric} & \textbf{Baseline} & \textbf{Robot} & \textbf{Robot+Fix.} & \textbf{$p$ (omni)} & \textbf{$Z$ (trend)} & \textbf{$p$ (trend)}\\
\midrule
%\multicolumn{7}{@{}l}{\emph{Traffic volume}}\\
Groups & 20{,}180 & 21{,}938 & 20{,}035 & -- & -- & -- \\
%\addlinespace[1pt]
%\multicolumn{7}{@{}l}{\emph{Traffic composition}}\\
\midrule
Child-only   & 2.72\%  & 28.96\% & 40.24\% & $<.001$ & --    & -- \\
Adult-only   & 76.19\% & 24.70\% & 17.91\% & $<.001$ & --    & -- \\
Child+Adult  & 20.41\% & 44.51\% & 41.85\% & $<.001$ & --    & -- \\
%\addlinespace[1pt]
%\multicolumn{7}{@{}l}{\emph{Funnel metrics}}\\
\midrule
Stopping                     & 0.73\%  & 1.50\%  & 2.48\%  & $<.001$ & 14.151 & $<.001$ \\
Sample touch                 & 47.62\% & 63.72\% & 46.68\% & $<.001$ & -1.951 & n.s. (.051) \\
Clerk approach               & 19.73\% & 13.41\% & 4.83\%  & $<.001$ & -5.850 & $<.001$ \\
Store entry                  & 31.97\% & 20.12\% & 9.46\%  & $<.001$ & -6.821 & $<.001$ \\
Staff-led entry              & 8.84\%  & 1.52\%  & 0.80\%  & $<.001$ & -4.999 & $<.001$ \\
Assisted in-store exp.       & 18.37\% & 6.10\%  & 4.02\%  & $<.001$ & -5.396 & $<.001$ \\
Purchase (per stopper)       & 11.56\% & 5.49\%  & 2.62\%  & $<.001$ & -4.314 & $<.001$ \\
Purchase (per passerby)      & 0.084\% & 0.082\% & 0.065\% & n.s. & -0.698 & n.s. \\
\bottomrule
\end{tabular}
\end{tighttable}
\begin{tablenotes}[flushleft]\footnotesize
\item Robot = Robot-only; Robot+Fix. = Robot+Fixture. \emph{$p$ (omni)} = omnibus Pearson’s $\chi^2$ across three conditions. Trend = Cochran–Armitage with ordering Baseline $<$ Robot $<$ Robot+Fix.\par
\end{tablenotes}
\end{threeparttable}
\end{table*}

\section{Findings}

We first overview quantitative shifts and then present qualitative themes that explain these shifts.

\begin{table*}[t]
\centering
\caption{Post hoc pairwise $\chi^2$ tests (Bonferroni-adjusted). Cell shows direction and $p$ for A$\to$B.}
\label{tab:quant_summaryposthoc}
\begin{threeparttable}
\begin{tighttable}
\begin{tabular}{@{}l c c c@{}}
\toprule
\textbf{Metric} & \textbf{Baseline$\to$Robot} & \textbf{Baseline$\to$Robot+Fix.} & \textbf{Robot$\to$Robot+Fix.}\\
\midrule
%\multicolumn{4}{@{}l}{\emph{Traffic composition}}\\
Child-only           & $\uparrow$, $p<.001$ & $\uparrow$, $p<.001$ & $\uparrow$, $p<.01$ \\
Adult-only           & $\downarrow$, $p<.001$ & $\downarrow$, $p<.001$ & n.s.\ ($p=.069$) \\
Child+Adult          & $\uparrow$, $p<.001$ & $\uparrow$, $p<.001$ & n.s. \\
%\addlinespace[1pt]
%\multicolumn{4}{@{}l}{\emph{Funnel metrics}}\\
\midrule
Stopping             & $\uparrow$, $p<.001$ & $\uparrow$, $p<.001$ & $\uparrow$, $p<.001$ \\
Sample touch         & $\uparrow$, $p<.01$  & n.s.                  & $\downarrow$, $p<.001$ \\
Clerk approach       & n.s.                  & $\downarrow$, $p<.001$ & $\downarrow$, $p<.001$ \\
Store entry          & $\downarrow$, $p<.05$ & $\downarrow$, $p<.001$ & $\downarrow$, $p<.001$ \\
Staff-led entry      & $\downarrow$, $p<.001$ & $\downarrow$, $p<.001$ & n.s. \\
Assisted in-store exp.& $\downarrow$, $p<.001$ & $\downarrow$, $p<.001$ & n.s. \\
Purchase (per stopper)& n.s.\ ($p=.094$)     & $\downarrow$, $p<.001$ & n.s. \\
\bottomrule
\end{tabular}
\end{tighttable}
\begin{tablenotes}[flushleft]\footnotesize
\item $\uparrow$ = increase from A to B (B $>$ A); $\downarrow$ = decrease (B $<$ A).\par
\end{tablenotes}
\end{threeparttable}
\end{table*}

\subsection{Quantitative Analysis}

We report video-annotation results across three conditions (Baseline, Robot-only, Robot+Fixture). Tables~\ref{tab:quant_summary} and \ref{tab:quant_summaryposthoc} summarize the results.
%Stopping is computed per passerby; all downstream outcomes are per stopper.

\textbf{Stopping.}
% \paragraph{\textbf{Stopping.}}
Passerby counts were 20{,}180 (Baseline), 21{,}938 (Robot-only), and 20{,}035 (Robot+Fixture). Stopping rates were 0.73\% (Baseline), 1.50\% (Robot-only), and 2.48\% (Robot+Fixture). A Pearson chi-square test of independence indicated a reliable effect of condition ($\chi^2(2)=201.55$, $p<.001$). Post hoc Bonferroni-adjusted pairwise tests showed Robot+Fixture $>$ Robot-only ($p<.001$) and Robot+Fixture $>$ Baseline ($p<.001$); Robot-only $>$ Baseline ($p<.001$). The entrance setup amplified attention as intended.

\textbf{Sample touch.}
% \paragraph{\textbf{Sample touch.}}
Sample touch denotes the proportion of stoppers who touched a sample product placed near the robot. Rates were 47.62\% (Baseline), 63.72\% (Robot-only), and 46.68\% (Robot+Fixture). The omnibus chi-square was significant ($\chi^2(2)=24.71$, $p<.001$). Pairwise comparisons showed Robot-only $>$ Baseline ($p<.01$) and Robot-only $>$ Robot+Fixture ($p<.001$).

\textbf{Clerk approach.}
% \paragraph{\textbf{Clerk approach.}}
Clerk approach is the proportion of stoppers who received a clerk-initiated approach. Rates were 19.73\% (Baseline), 13.41\% (Robot-only), and 4.83\% (Robot+Fixture). The omnibus test was significant ($\chi^2(2)=34.54$, $p<.001$). Post hoc tests indicated Baseline $>$ Robot+Fixture ($p<.001$) and Robot-only $>$ Robot+Fixture ($p<.001$); the Baseline vs.\ Robot-only contrast was not significant.

\textbf{Store entry.}
% \paragraph{\textbf{Store entry.}}
Store entry is the proportion of stoppers who crossed the threshold into the interior. Rates were 31.97\% (Baseline), 20.12\% (Robot-only), and 9.46\% (Robot+Fixture). The omnibus test was significant ($\chi^2(2)=46.65$, $p<.001$). Pairwise: Baseline $>$ Robot-only ($p<.05$), Baseline $>$ Robot+Fixture ($p<.001$), and Robot-only $>$ Robot+Fixture ($p<.001$).

\textbf{Staff-led entry.}
% \paragraph{\textbf{Staff-led entry.}}
Staff-led entry is the proportion of stoppers who entered immediately after a clerk approach. Rates were 8.84\% (Baseline), 1.52\% (Robot-only), and 0.80\% (Robot+Fixture). The omnibus test was significant ($\chi^2(2)=34.36$, $p<.001$). Pairwise: Baseline $>$ Robot-only ($p<.001$) and Baseline $>$ Robot+Fixture ($p<.001$).

\textbf{Assisted in-store experience.}
% \paragraph{\textbf{Assisted in-store experience.}}
Assisted in-store experience denotes guided tryouts or explanations ($>$30\,s) per stopper. Rates were 18.37\% (Baseline), 6.10\% (Robot-only), and 4.02\% (Robot+Fixture). The omnibus test was significant ($\chi^2(2)=36.85$, $p<.001$). Pairwise: Baseline $>$ Robot-only ($p<.001$) and Baseline $>$ Robot+Fixture ($p<.001$).

\textbf{Purchase.}
% \paragraph{\textbf{Purchase.}}
Purchase is the proportion of stoppers who made a purchase. Rates were 11.56\% (Baseline), 5.49\% (Robot-only), and 2.62\% (Robot+Fixture). The omnibus test was significant ($\chi^2(2)=19.67$, $p<.001$). Post hoc tests showed Baseline $>$ Robot+Fixture ($p<.001$); the other contrasts were not significant after adjustment.

\textbf{Purchase (per passerby).}
% \paragraph{\textbf{Purchase (per passerby).}}
For completeness, we evaluated Purchase (per passerby): the proportion of passersby who made a purchase. Rates were 0.084\% (Baseline: 17/20{,}180), 0.082\% (Robot-only: 18/21{,}938), and 0.065\% (Robot+Fixture: 13/20{,}035). The omnibus chi-square was not significant ($\chi^2(2)=0.59$, $p=.744$).

\subsection{Qualitative Analysis}

We analyzed six post-experiment clerk interviews (one manager P1, five clerks P2--P6) using reflexive thematic analysis grounded in the coding table and field notes. Theme counts below indicate coded utterances (occurrences). We report five themes with representative, de-identified quotes (translated) and explicit links to the quantitative funnel.

\paragraph{\textbf{Theme overview.}}

Theme counts (utterances) were: T1~attraction $n{=}30$ (positive $n{=}20$, negative $n{=}10$), T2~conversion barriers $n{=}8$, T3~timing/priority $n{=}18$, T4~role positioning $n{=}21$, T5~fixture design/effect $n{=}29$ (design $n{=}5$, effect $n{=}24$). These frequencies are descriptive rather than effect sizes. All six participants contributed to at least three themes.

\paragraph{\textbf{T1. Entrance anchoring raises stopping but limits progression.}}

Clerks consistently reported that the entrance-placed robot acted as a strong visual anchor that drew gaze and prompted passersby to pause. The attractor effect was especially pronounced among children, which in turn increased the share of family groups who stopped compared to the Baseline (no robot) condition.

\begin{quote}
``\ldots the draw naturally skewed toward small children, but if that interest brings people a step closer to the storefront it creates traffic; from the store’s perspective it clearly increases visit opportunities and is a large benefit.'' [P3, translated]
\end{quote}
\begin{quote}
``Without the robot, families with children don’t really stop there---our shoppers are mainly adults.'' [P2, translated]
\end{quote}

At the same time, clerks described difficulty converting that attention into downstream steps in the service funnel. Many parents (adult decision-makers) accompanied children without a strong shopping intention, and some customers were attracted by the robot itself rather than the products---they chatted briefly, felt satisfied, and left.

%\begin{quote}
%``Fathers (and/or mothers) often come only as companions; they may or may not have any real product interest.'' [P2, translated]
%\end{quote}
\begin{quote}
``Children were the ones most drawn to the robot, but children don’t buy beds; getting from there to a purchase meant a different negotiation route with the parents. If this were toys or children’s learning products it would be a perfect fit. [unclear]'' [P5, translated]
\end{quote}

This qualitative pattern aligns with our quantitative results: \emph{Stopping} per passerby rose under Robot-only and Robot+Fixture, while \emph{Store entry} per stopper declined in both conditions.

\paragraph{\textbf{T2. Conversion barriers for high-ticket items and family groups.}}

Clerks questioned whether the current deployment could carry customers through evaluation and closing for high-ticket goods. Price point and conversational smoothness were repeatedly cited as hurdles.

\begin{quote}
``It’s not a trivial purchase; these are high-ticket items, so smooth communication matters. If it were a hundred or two hundred yen, fine---but here we’re talking four to five hundred thousand yen.'' [P1, translated]
\end{quote}
\begin{quote}
``Because these items are expensive, the initial barrier to coming in is high. [unclear]'' [P5, translated]
\end{quote}

As in T1, clerks noted that attraction centered on children does not readily translate into adult evaluation or purchase. Interest in the robot and interest in the products are different things; several participants described interactions that ended at light chat without progressing.

%\begin{quote}
%``Honestly I couldn’t tell much; it ended up being mostly conversations with children, which didn’t lead to service. I wished we could have developed those interactions more---that was the weak point this time.'' [P2, translated]
%\end{quote}
\begin{quote}
``It was easy to start conversations---very easy---but it inevitably centered on kids. The talk always flows toward pillows. From there you can push a bit---`We recommend this,' `Kids can use it too'---but very few cases led to a purchase, so it was hard.'' [P6, translated]
\end{quote}

One participant put it bluntly:

\begin{quote}
``We don’t actually want child traffic\ldots'' [P1, translated]
\end{quote}

Together, these accounts help explain the quantitative pattern: despite increased \emph{Stopping} per passerby, downstream, clerk-led steps were lower in robot-present conditions---notably reduced \emph{Clerk approach}, \emph{Staff-led entry}, \emph{Assisted in-store experience}, and \emph{Purchase}. Clerks’ motivation to actively approach and provide service decreased when they realized that many customers were not actually interested in purchasing.

\paragraph{\textbf{T3. Non-intrusion timing and priority delay human approach.}}

Clerks often felt unsure about when to approach customers. It was hard to read the state of the robot--customer exchange and to find a suitable moment to join the conversation. Latency in response generation and occasional speech recognition errors further complicated the timing.

\begin{quote}
``We have to watch for a pause to find the right timing to approach; if we could remove that constraint it would be better. If we speak while it’s speaking, customers won’t listen; even while the robot is talking, we’d like the customer’s voice to be prioritized so we can transition into a tryout.'' [P1, translated]
\end{quote}
\begin{quote}
``When we asked questions, Sota sometimes had a thinking pause or misheard us even when we spoke clearly.'' [P2, translated]
\end{quote}

To avoid interrupting, clerks sometimes chose to wait while customers talked with the robot---especially when children were involved. This waiting reduced the likelihood of initiating service, which in turn lowered clerks’ motivation to approach.

\begin{quote}
``I tried to let them enjoy it a bit and waited; once the child was satisfied, I stepped in.'' [P2, translated]
\end{quote}
%\begin{quote}
%``Typically customers start by touching; I wait---if they keep touching for a while it’s hard to cut in. If they’re just pat-patting, I wait a bit, about ten seconds; I decide based on dwell time.'' [P6, translated]
%\end{quote}

Some clerks developed new approach patterns distinct from their usual routine, coordinating their entry at a visible pause or via a family handoff.

\begin{quote}
``With families, the child engages first and calls the parent over; at that timing we can coordinate a handoff---`Would you like to try it, ma’am?' '' [P4, translated]
\end{quote}

These phenomenons helps explain the quantitative pattern: despite higher \emph{Stopping} at the entrance, robot-present conditions showed lower \emph{Clerk approach}, \emph{Staff-led entry}, and \emph{Assisted in-store experience}.

\paragraph{\textbf{T4. Role positioning: robot as attractor/PR, staff as closer.}}

Across interviews, clerks consistently positioned the robot as a front-of-store attractor or ``PR support,'' while humans handled evaluation, recommendation, and closing. In this division of labor, the robot’s job is to spark interest and create openings; staff then step in to guide product trial and decision-making.

\begin{quote}
``I see the robot’s role as getting people to visit and become interested in the store; once they come inside, we do the explaining, and each staff member’s way of explaining differs.'' [P4, translated]
\end{quote}
%\begin{quote}
%``For attracting people it worked extremely well---its job is to spark interest.'' [P6, translated]
%\end{quote}
\begin{quote}
``\ldots positioning it as PR support rather than the frontline could work well.'' [P1, translated]
\end{quote}

Participants also noted that the robot can lightly say things or use phrasing that clerks might avoid (e.g., local dialect), which can prime interest and convey friendliness, though tone management might be important in a higher-end retail context.

\begin{quote}
``Using Kansai dialect created friendliness, but stricter customers might find it impolite.'' [P6, translated]
\end{quote}
\begin{quote}
``Because it is a robot, it can say things we would not; a bit of humor fits here.'' [P3, translated]
\end{quote}

\paragraph{\textbf{T5. Staging and affordances shape approachability.}}

Clerks generally praised visual staging (using a fixture) together with immediate product access at the entrance (e.g., touch-and-experience) for making the setup more noticeable and inviting.

\begin{quote}
``A robot on site already attracts interest; we discussed lowering it, but I prefer larger/taller---more chances to be seen, a stronger impression, and a routine of `see robot -> look at product'.'' [P1, translated]
\end{quote}
%\begin{quote}
%``The large, heavy fixture we used this time was eye-catching; between versions, I personally prefer having the fixture if Sota is going to stand there.'' [P3, translated]
%\end{quote}
\begin{quote}
``When it was there, even from far away customers would say, `Huh, what's that?'---it catches the eye at a distance.'' [P4, translated]
\end{quote}

Participants also offered concrete fixture-design suggestions. Height and placement should prioritize adult approachability while maintaining a clear `territory' around the robot that discourages overly close child interaction and protects the device.

\begin{quote}
``Adults converse at that height---you might lower it a little, but with average male height around 170\,cm, setting 150--160\,cm lets adults look down slightly and speak comfortably; if it's too low, kids will be the ones to engage.'' [P1, translated]
\end{quote}
\begin{quote}
``Without a fixture the robot feels closer, which made kids try to touch it more---we hesitated about stopping them; the fixture created a sense of territory for Sota, and fewer kids touched the robot instead of the products.'' [P5, translated]
\end{quote}

Together, these remarks offer pragmatic guidance for entrance staging: use fixtures to boost visibility, place product samples within immediate reach, and tune height/placement to preserve adult approachability while reducing inadvertent child-dominated interaction.

\section{Discussion}
\subsection{Summary of Findings}

\begin{figure*}[t]
\centering
\resizebox{\textwidth}{!}{%
\begin{tikzpicture}[
  node distance=8mm and 7mm,
  box/.style={
    rectangle, rounded corners, draw=black, very thick,
    align=center, text width=24mm, %
    minimum height=8mm, inner sep=3pt
  },
  gate/.style={
    rectangle, draw=black, very thick, dashed,
    align=center, text width=24mm, %
    minimum height=8mm, inner sep=3pt
  },
  note/.style={font=\small, align=center},
  >={Stealth[length=2.2mm]}
]
% Nodes
\node[box] (pass) {Passersby};
\node[box, right=of pass] (stop) {Attract\\(Stop)};
\node[gate, right=of stop] (handoff) {Handoff\\(Robot -> Clerk)};
\node[box, right=of handoff] (approach) {Approach};
\node[box, right=of approach] (entry) {Entry};
\node[box, right=of entry] (assist) {Assisted};
\node[box, right=of assist] (purchase) {Purchase};

% Main arrows
\draw[->, very thick] (pass) -- (stop);
\draw[->, very thick] (stop) -- (handoff);
\draw[->, very thick] (handoff) -- (approach) -- (entry) -- (assist) -- (purchase);

\node[box, below=14mm of handoff, text width=28mm] (exit) {Early Exit / Drift};
%\draw[->, thick, dashed, shorten >=2pt]
%  (stop.south) to[out=-90, in=90] (exit.north);
%\draw[->, thick, dashed, shorten >=2pt]
%  (stop.south east) to[out=-60, in=90] (exit.north);
\draw[->, thick, dashed, shorten >=2pt]
  (stop.south) to[out=-90, in=90] (exit.north);

% Role labels
\node[note, above=2mm of stop] {\footnotesize Robot-led};
\node[note, above=2mm of approach] {\footnotesize Clerk-led};

% Annotation
\node[note, below=2mm of handoff] {\footnotesize Ambiguous cues \& non-intrusion timing => lower clerk-led steps per stopper};

\end{tikzpicture}%
} % end resizebox
\caption{Integrative chain. \emph{Attract} (robot-led) raises stopping at the entrance; progression depends on a clear \emph{Handoff} to clerk-led evaluation. Without it, many encounters \emph{close at the entrance}. \textbf{Metric mapping:} Stopping per passerby -> Attract; Sample touch -> threshold tryability; Clerk approach / Store entry / Assisted -> Handoff \& Interior; Purchase per stopper (terminal), Purchase per passerby (overall volume).}
\label{fig:attract-handoff-interior}
\end{figure*}
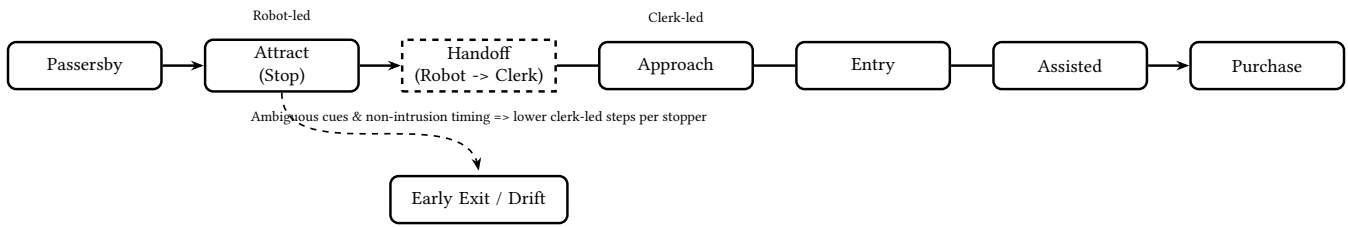

In a 12-day field deployment with three conditions (Baseline, Robot-only, Robot+Fixture), video annotations revealed a consistent pattern: the robot increased Stopping per passerby (largest with the fixture), yet clerk-led downstream steps per stopper--clerk approach, store entry, assisted experience, and purchase--declined; purchases per passerby did not change reliably. 

Staff interviews help explain this divergence: attraction was often child-centered with low immediate intent for high-priced items, while non-intrusion norms and occasional conversational latency made clerks wait rather than interrupt, dampening approach motivation and yielding missed handoffs from entrance interest to interior evaluation. Note that Stopping is computed per passerby; all downstream metrics are per stopper to isolate post-attraction effects. Because these are ratios, a condition can show lower rates even when absolute counts are similar or higher; we therefore foreground ratio-based analyses to characterize the robot’s influence conditional on stopping.

\subsection{Attract => Handoff => Interior: an integrative account}

Our results can be summarized by a simple chain (Fig.~\ref{fig:attract-handoff-interior}): the entrance robot reliably increases Attract (higher stopping per passerby); progression then hinges on a clear, lightweight \emph{Handoff} from robot-led talk to human-led evaluation; finally, interior Evaluation/Service (approach, entry, assisted experience, purchase) materializes when the handoff is observable and timely.

In our deployment, many encounters appeared to close at the entrance. For customers without strong purchase intent, a quick Q\&A plus a brief touch-try often satisfied immediate curiosity and minimal product questions. This ``closure of needs at the robot'' was more pronounced with the fixture, which created a well-defined micro-space: it improved visibility and tryability, yet risked becoming perceptually separate from the store interior. Quantitatively, this aligns with a monotonic decrease from Baseline to Robot+Fixture in clerk-led steps per stopper (approach, entry, assisted experience, purchase), despite stopping increasing with the robot present.

A practical implication is to treat the handoff as first-class: agree on lightweight triggers (e.g., a short lull, a simple robot cue, or a ``touch > $N$ s'' heuristic), make the handoff observable to both parties (e.g., a brief utterance like ``A clerk will assist you now''), and preserve sightlines so clerks can time their first turn without hovering. These choices strengthen continuity from the threshold to the interior while retaining the robot’s attention gains.

\subsection{Design tensions: Visibility vs.\ approachability}

We included the fixture as a pragmatic, visually oriented staging choice agreed with the store. Clerks generally endorsed using the robot and fixture as a set and pointed to concrete parameters that matter in practice: size (visual impact vs.\ obstruction), platform height (adult-comfortable talk while discouraging child-only play), color/materials (brand fit), movability (layout changes), and finish/robustness (safety and upkeep).
% We included the fixture as a pragmatic, visually oriented staging choice agreed with the store, not as a theoretical focus. 

Quantitatively, the fixture amplified visibility and stopping, yet downstream clerk-led steps (approach, entry) fell. Interviews suggest a mechanism: a well-defined “stage” at the threshold helps people notice and try the sample but can also become a separate micro-space where curiosity is satisfied and progression stalls. In short, higher visibility does not guarantee approachability.

A key design implication is to balance entrance salience with clear paths inward. Preserve sightlines so staff can monitor without hovering; avoid fully bounded pedestals; favor lighter or transparent elements; place touchable samples slightly inside the threshold to invite forward motion; set the interaction height for adults in high-end contexts; and keep the unit movable so staging can be tuned to traffic and staff coverage. These choices retain attention gains while reducing missed handoffs.

\subsection{Implications for triadic coordination}

Retail robot deployment is not a dyad between robot and customers; it is a socio-technical arrangement involving robot, clerks, customers, managers, and, at times, product and system designers. Effective frontline service therefore depends on triadic coordination in which robots and clerks act in complementary roles, and managers provide policies and resourcing that make those roles viable. The handoff from robot-led attraction to human-led evaluation is the bridge that ties the triad together.

We summarize actionable design implications for researchers and industry practitioners:

\textbf{Clarify roles and triggers.}
% \paragraph{\textbf{Clarify roles and triggers.}}
Define who does what and when. A clear and legible purpose for the robot is essential, as unclear purpose can be a pitfall in retail deployments and dampens conversation \cite{OkafujiDesign2025}. Make explicit the robot’s attractor role and the clerk’s evaluator/closer role, and agree on concrete handoff triggers (e.g., a conversation lull, customer touch exceeding a set duration, or a robot cue). Keep the routine lightweight so it fits daily operations.

\textbf{Make handoffs observable.}
% \paragraph{\textbf{Make handoffs observable.}}
Provide simple cues that signal transition: a short robot utterance that invites staff (``A clerk will assist you now.''), a brief light/screen cue oriented toward staff, or a subtle gaze/body cue toward the interior. Place the robot where clerks can see and hear the interaction without hovering. We notice that this policy may be controversial, as prior work reports that explicitly calling a clerk can elicit negative customer attitudes \cite{SongFrom2025}. We view this tension as precisely why a mixed-methods design is valuable: it surfaces boundary conditions and practical cues for appropriate handoff design. Researchers should treat handoff signaling as context-dependent and tune the strength and modality of the cue (e.g., brief utterance vs.\ light/gaze cue), monitor reactions, and adapt during deployment in real environments.

\textbf{Share minimal state.}
% \paragraph{\textbf{Share minimal state.}}
Equip the system to surface a one-sentence summary of the ongoing topic (e.g., ``Discussing pillow firmness and size'') on a small staff-facing display or mobile view. This reduces approach anxiety and speeds the clerk’s first turn without requiring long transcripts or heavy tools. Keep summaries staff-only and avoid storing personally identifiable information.

\textbf{Balance visibility with approachability.}
% \paragraph{\textbf{Balance visibility with approachability.}}
Design the entrance staging (with or without a fixture) so that the natural next step is toward the interior and staff. Preserve sightlines, avoid creating a fully separate ``stage,'' and locate touchable samples slightly inside the threshold to encourage forward motion rather than lingering.

\textbf{Support discretion without ambiguity.}
% \paragraph{\textbf{Support discretion without ambiguity.}}
Provide short training on when to interrupt and how to phrase the first turn; supply example scripts and exceptions for peak hours. Managers should set priorities when duties compete (e.g., who covers handoffs during checkout rush) so clerks are not penalized for timely approaches.

\textbf{Align metrics and governance.}
% \paragraph{\textbf{Align metrics and governance.}}
Track not only stopping but also handoff success, interior evaluation starts, and assisted experiences. Hold brief retros with staff to tune triggers, scripts, and layout. Align incentives so clerks are recognized for effective handoffs, not just for final sales, to avoid perverse disincentives.

\textbf{Transferability by involvement level.}
Despite the single-site scope, we contend that the findings transfer to retail settings with similar demands for clerk hospitality. Transferability is particularly plausible for high-involvement goods, where current social robots may be insufficient to meet service requirements and clerk expertise is highly valued; in such cases, the robot–customer–clerk configuration can be effectively leveraged.
% By contrast, for low-involvement goods that do not require intensive clerk hospitality, a robot alone may be sufficient as an entrance attractor.

\subsection{\textbf{Limitations and Future Work.}}

We have several limitations in our study. First, we deployed our robot only at the entrance, not in other locations (e.g., inside the store). As a result, clerks were not able to experience different settings and compare them to reach a more holistic view of the robot. We suggest that future work expand this study and explore different deployment patterns to gain deeper insights into the retail robot’s role and benefits.

%Second, we conducted post-experiment interviews before the full video annotation and analysis, in order to preserve recall and due to practical operational limitations with the store. Future work may experimentally manipulate system mechanisms such as handoff, model day/time effects with mixed models, and explore longer-term performance changes for both clerks and customers.

Second, we consider longer exposure may moderate both attention and conversion. As customers and staff acclimate to the robot and its entrance staging, novelty-driven stopping could decline while clerk timing and handoffs become more efficient; the overall balance between Attract and Interior may thus shift. We therefore encourage future work to pursue longitudinal deployments that track adaptation on both sides.

Finally, our observations may largely depend on the appearance of the Sota robot. Researchers in the future should expand studies to a variety of robot types, including androids and virtual CG agents.

\section{Acknowledgements}
%\begin{acks}
%AI tools, such as ChatGPT, are used in the writing of the paper for translation and organization of the texts. However, the authors declare that we originate and create the idea and work.
This work was supported by JST Moonshot R\&D Grant Number JP-MJMS2011.
%\end{acks}

\bibliographystyle{ACM-Reference-Format}
\bibliography{sample-base}

\end{document}